\documentclass[letterpaper, 10 pt, conference]{ieeeconf}
\IEEEoverridecommandlockouts
\usepackage{cite}
\usepackage{amsmath,amssymb,amsfonts}
\usepackage{algorithmic}
\usepackage{graphicx}
\usepackage{textcomp}
\usepackage{xcolor}
\usepackage{float}

\title{ \LARGE \bf
General Hand Guidance Framework using Microsoft HoloLens
}

\author{David Puljiz$^{1}$,  Erik St\"ohr$^{1}$, Katharina S. Riesterer$^{1}$, Bj\"orn Hein$^{1,2}$, Torsten Kr\"oger$^{1}$
\thanks{$^{1}$Intelligent Process Automation and Robotics Lab, Karlsruhe Institute of Technology, Karlsruhe, Germany {\tt\small david.puljiz@kit.edu, ujeji@student.kit.edu, uxecu@student.kit.edu, bjoern.hein@kit.edu, torsten@kit.edu}}%
\thanks{$^{2}$Karlsruhe University of Applied Sciences, Karlsruhe, Germany}%
}

\begin{document}
\maketitle
\thispagestyle{empty}
\pagestyle{empty}

\begin{abstract}
Hand guidance emerged from the safety requirements for collaborative robots, namely possessing joint-torque sensors. Since then it has proven to be a powerful tool for easy trajectory programming, allowing lay-users to reprogram robots intuitively. Going beyond, a robot can learn tasks by user demonstrations through kinesthetic teaching, enabling robots to generalise tasks and further reducing the need for reprogramming. However, hand guidance is still mostly relegated to collaborative robots. Here we propose a method that doesn't require any sensors on the robot or in the robot cell, by using a Microsoft HoloLens augmented reality head mounted display. We reference the robot using a registration algorithm to match the robot model to the spatial mesh. The in-built hand tracking and localisation capabilities are then used to calculate the position of the hands relative to the robot. By decomposing the hand movements into orthogonal rotations and propagating it down through the kinematic chain, we achieve a generalised hand guidance without the need to build a dynamic model of the robot itself. We tested our approach on a commonly used industrial manipulator, the KUKA KR-5. \par  

\end{abstract}


\section{Introduction}

With the emergence of collaborative robots \cite{bischoff2010kuka}, such as the Universal Robotics' UR series, KUKA's LBR iiwa and others, hand guidance has become ubiquitous. This is due to the safety requirements needing joint-torque sensors (JTS) or similar sensors on joints. This in turn allowed the gravity compensation mode, that is hand guidance, which allows easy teaching of new trajectories, even by lay-users. Such ease of teaching and reprogramming is of extreme importance to expand robotics to small and medium enterprises (SME) as well as in flexible manufacturing paradigms \cite{schraft2006need}. Hand guidance on industrial robots in the mean time has been marginal, usually requiring external force-torque (FTS) or other sensors in the robot cell. Traditional teaching pedants are still the most used tool to program industrial robots, despite requiring extensive training to use and being slow when it comes to reprogramming. Besides industrial settings, hand guiding is also important in medical robotics \cite{Hanses2016predefined}. \par 

Beyond simply easing trajectory teaching, hand guidance is an essential part of Kinesthetic learning, a subfield of Imitation Learning (also called learning from demonstration -LfD) \cite{argall2009survey}. The goal of imitation learning is to teach robot skills purely from human actions. This can be done either by observation and tracking of human movements and actions and then adapting them to a robot, or by a human directly moving the robot, either real or virtual, to demonstrate how they would perform the action. The later is comprised of kinesthetic teaching and teaching through teleoperation. Imitation learning can greatly reduce the search space in learning new tasks, as it already receives positive examples from demonstrations.  \par

The use of Augmented Reality (AR) in the field of robotics is a long yet sporadic research field. Due to the lack of practicality and robustness in early head-mounted devices (HMDs), most of the earlier research used either camera and screen \cite{bischoff2004perspectives} or a projector \cite{gaschler2014intuitive} to display information. The release of the Microsoft HoloLens AR-device in 2016 marked the emergence of practical HMDs, allowing more flexible applications not bounded to an already set-up environment. Wearable AR has been used to ease robot programming \cite{Quintero2018ARProg}, knowledge patching for imitation learning \cite{liu2018interactive}, task planning for collaborative human-robot workspaces \cite{chakraborti2017} and for the display robot information to humans \cite{Walker2018robotmotion}, among other things. \par

\begin{figure}[t]
    \includegraphics[width=0.48\textwidth]{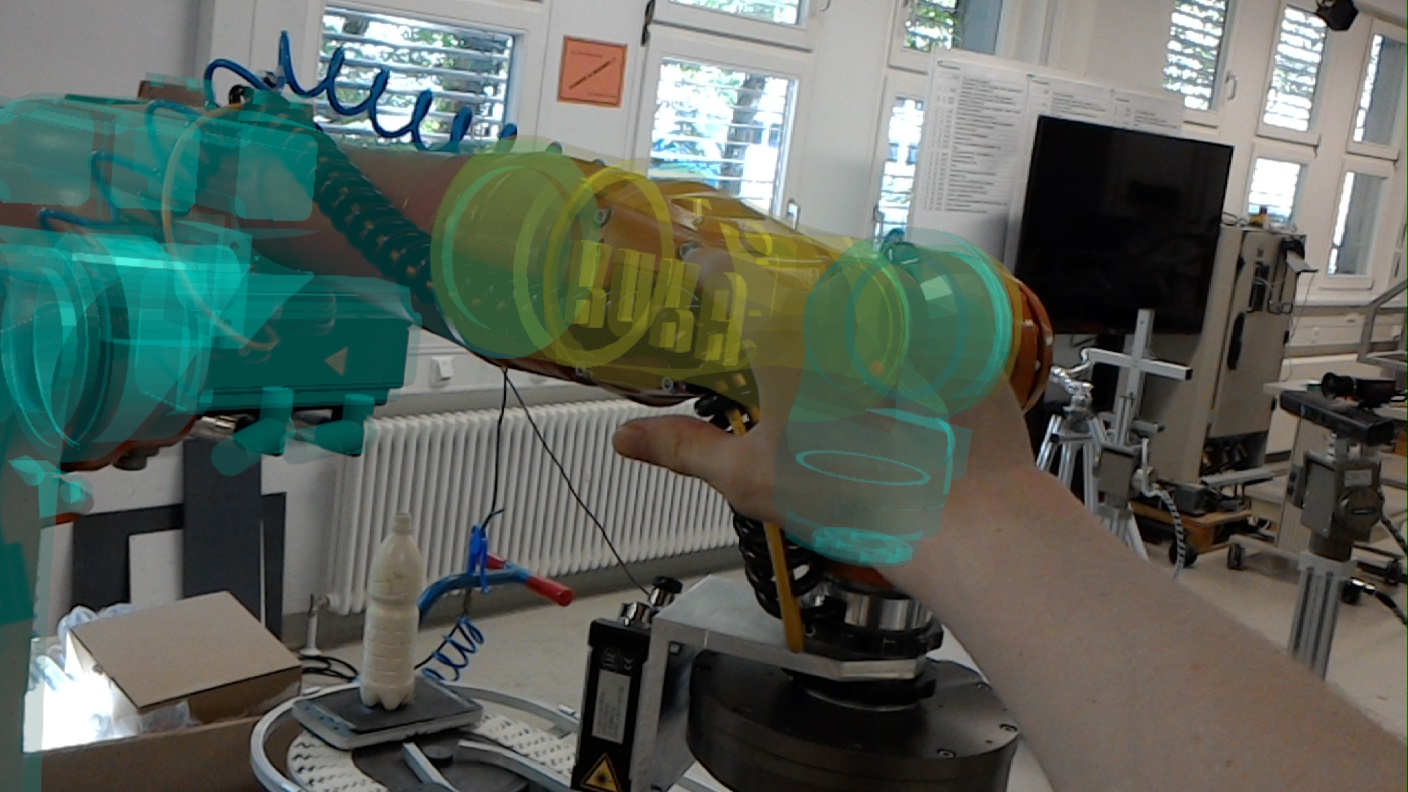}
    \caption{The user moving their hand next to the robot link they want to move. Note that the link turns yellow to indicate which link will be moved}
    \label{fig:intro}
\end{figure}

\subsection{Related work}

Methods of implementing hand guidance without joint sensors already exist. Moe \textit{et al.} use a Microsoft Kinect and a smartphone-based accelerometer to perform hand tracking and guide the end-effector of an industrial robot \cite{Moe2013hg}. This approach however was limited to 5DOF. Furthermore by just driving the end-effector one can not make use of extra degrees of freedom that robots with seven or more joints provides. \par

Lee \textit{et al.} proposed a generalised method of hand guidance by torque control, based on the dynamic model of a robot, the motor current, and the joint friction model \cite{Lee2016sensorless}. The approach, however, requires experiments to determine the friction model of each joint, as well as possessing a dynamic model. Furthermore the robot is confined to using a torque controller. Finally the external force needed to move the end-effector was found to be 1.23-4.83 times greater than approaches based on JTS. Ideally the force would be close to zero, which isn't the case even for JTS methods. \par

A very similar approach was taken by Ragaglia \textit{et al.} \cite{RAGAGLIA2016AccurateSensorless}, requiring a dynamic model and friction model identification, using a voting system and admittance control to move the robot. These two factors, as with the previous approach, prevent it to be a truly "plug-and-play" method. \par

In \cite{Stolt2015Sensorless} a control strategy was proposed without joint friction model identification, although a feed-forward term for partial friction compensation could be added. It was noticed that with such a method the force required to move larger robots becomes too large for human operators, thus the approach is not general. Likewise it has a fixed control architecture that cannot be changed. \par

In this paper we propose an easily transportable method based on wearable sensors, thus not requiring any sensors on or around the robot. Our method requires only a kinematic model and works out of the box, not requiring any experiments or special setup before the use. Furthermore it can work with any controller, the external force needed to move the robot is zero as no force needs to be directly applied to the robot, and the sensitivity of the movement can be swiftly modified on-line. Finally it can be easily coupled with other wearable-AR-based robot programming and visualisation modalities. Particular interaction can be seen with \cite{liu2018interactive} allowing both Kinesthetic teaching and knowledge patching on a single device. \par 

\begin{figure*}[]
\centering

    \includegraphics[width=0.4\textwidth]{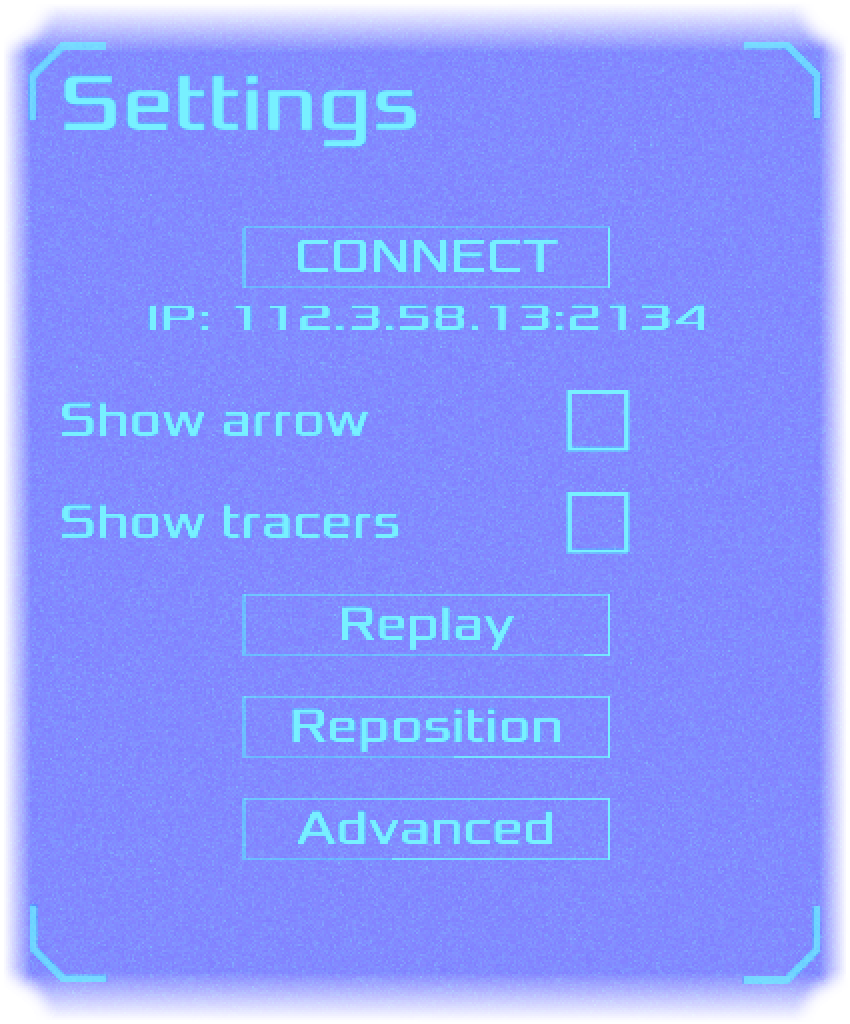}
    \includegraphics[width=0.4\textwidth]{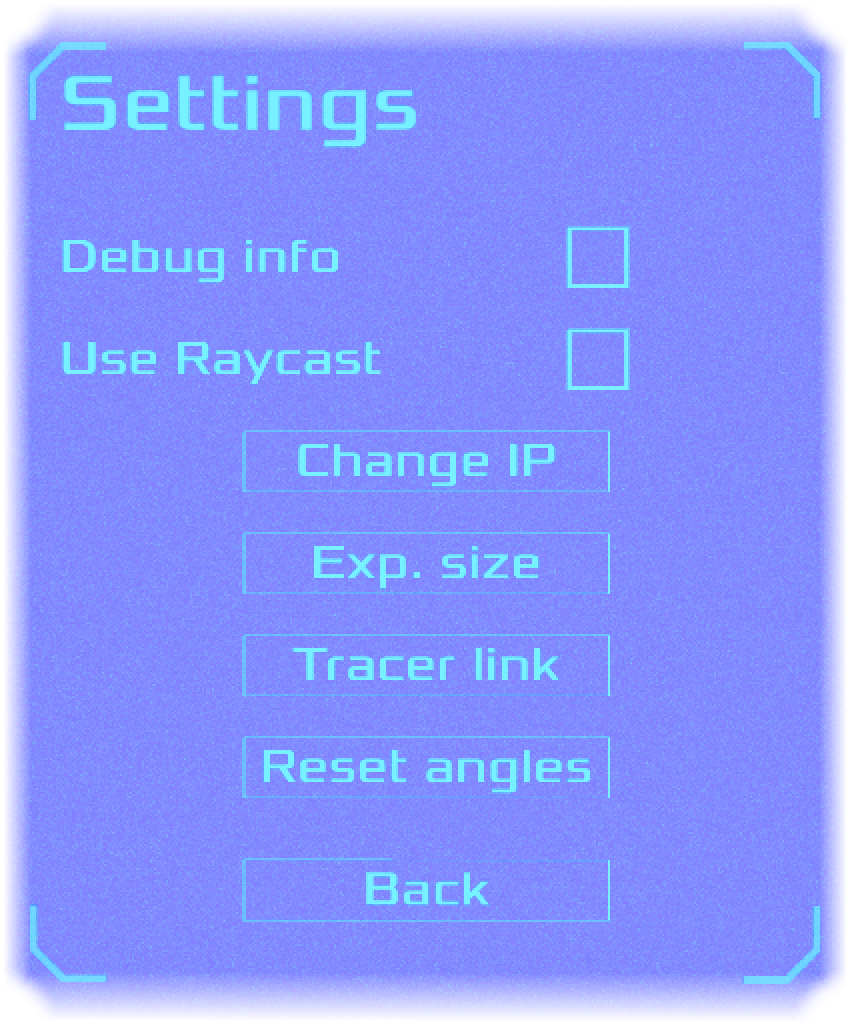}
    \includegraphics[width=0.85\textwidth]{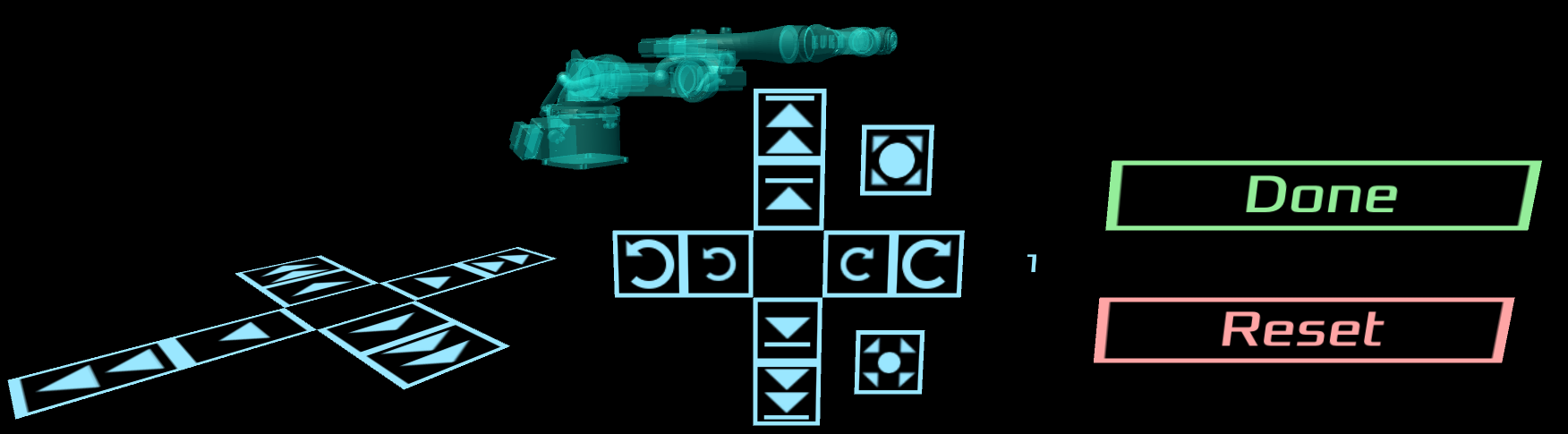}
    \caption[]{The UI of the application. On the top left one can see the main menu following the user. It allows to 1) open connections with a PC running the required ROS nodes 2) show arrows or tracers that illustrate better the hand movements detected 3) replay the trajectory performed 3)reposition and resize the robot (illustrated on the bottom picture) 4) open the advanced menu shown on the top left; The advanced menu is used for debugging, changing the IP and expanding the size of the space in which the hand movement activates robot movements; On the bottom options for repositioning and resizing the robot are shown.  }
     \label{fig:interface}
\end{figure*}

\subsection{Contributions}

This paper presents three main contributions to the state of the art:
\begin{enumerate}
    \item We develop a freely tuneable, flexible system for hand guidance, which doesn't need any sensors on or around the robot, any special identification experiments before use or any specific controllers. It works with any type of robot, and, if the robot is inaccessible, it allows teleoperation.
    \item We incorporate and test a semi-autonomous robot to headset referencing system that successfully overlays the virtual robot over the real one and also tracks the position of the headset in reference to the robot.
    \item We integrate these interaction modalities on a widely use AR device, using open-source tools. This allows the solution to be easily compatible with other AR interaction modalities and programs, thus increasing the scope of AR in human-robot interaction (HRI) systems 
\end{enumerate}

The structure of this paper is as follows. The next section presents the system overview and the referencing method we used to overlay the digital robot onto the physical one. In Section \ref{sec:control} we describe how we generate desired joint states from hand motions and the control architecture. Section \ref{sec:exp} describes the experiments we conducted and the results. Finally, conclusions are drawn and the future work is outlined in Section \ref{sec:conc}. \par

\section{System and Referencing}
\label{sec:sys_ref}

The system consists of three main parts - the HoloLens, a computer and the KR-5 robot. The computer is running the open-source Robot Operating System (ROS) \cite{ros}, and is connected to the KR5. The workflow is as follows - the AR device is started and connects to the computer running ROS, which sends it the Universal robot description file (urdf) and associated visual and collision meshes. This allows the device to generate a holographic robot on-line with the same geometry and kinematics as the real one. \par

The user can then choose to place the robot manually or to use the semi-automatic referencing. The manual mode likely won't result in a good match between the real robot and the hologram, and is intended for teleoperation scenarios where the robot is either inaccessible or when it's hard to manually reach particular links due to the size of the robot. In the latter case where the robot is too large the hologram can also be resized. \par

The semi-automatic mode requires the user to place a "seed" hologram near the robot base and orient it approximately to the front of the robot. A registration algorithm is then used to refine the rough guess of the user, both increasing the precision and reducing the user strain of a purely manual method. The algorithm is described in detail in Section \ref{sec:ref}. \par

Using the in-built hand-tracking and localisation capabilities of the HoloLens, we detect when the hand is near the robot by identifying if the hand is inside a resized convex collision mesh, which is generated from the meshes received from the computer. If the hand is inside this active zone, the motions of the hand are converted to the desired joint values of the robot. ROS Control \cite{ros_control} is then used to select the desired controller. These steps are described in Section \ref{sec:control}. Desired trajectories can be saved and replayed through a menu on the HoloLens itself. The UI can be seen in Fig.~\ref{fig:interface} \par

\subsection{Referencing Algorithm}
\label{sec:ref}

The user first positions the seed hologram, in this case a cube, near the base of the robot and orients its z-axis(the axis is also visualised) to point approximately out of the front of the robot. We assume that the robot is static at the moment this is done and that the joint values are known to the controlling computer which in turn sends them to the AR device. The seed serves a dual purpose. Firstly it limits the search space of the registration algorithm, thus making convergence faster as well as preventing the registration to get stuck in a local minimum. Secondly it defines the centre of the area of the spatial mesh to be extracted. The spatial mesh is automatically generated by the HoloLens. We extract all the mesh in 2.5 meter radius of the seed hologram. The size can automatically be determined by the bounding box of the robot itself. The mesh is then sampled, converted to a point cloud and sent to the computer. Both the number of triangles per cubic meter and the sampling radius can be controlled resulting in larger or smaller point clouds. On the side of the computer, the cloud is filtered using the Moving Least Squares Method (MLS) and then either a standard Iterative Closest Point (ICP) algorithm \cite{icp} or the Super4PCS \cite{super4pcs} is used to refine the users initial guess by registering the model to the scene point cloud. This was implemented using the Point Cloud Library \cite{Rusu_ICRA2011_PCL}. The evaluation of the approach is described in Section \ref{sec:exp}. \par   

\section{Joint Command Calculation and Control}
\label{sec:control}

Once a successful overlay is achieved, a resized convex meshes around the virtual (and now also real) robot links define the area in which hand movements translate to robot movements. The size of the area can be freely modified online through a holographic menu. A pre-programmed \textit{Hold} gesture, defined in the HoloLens' specifications, is used to virtually push and pull the robot links. \par

When the user makes the \textit{Hold} gesture inside the mesh of link \textit{j} whose parent joint at position $s_{j,t}$ has a rotation axis $a_{j,t}$, the positions of the hand in the current frame $h_{t}$ and the position of the hand in the previous frame $h_{t-1}$ are projected unto the plane defined by $s_{j,t}$ and $a_{j,t}$ as per \eqref{eq:proj}:\par

\begin{equation}\label{eq:proj}
p_{j,t}=h_{t}-\frac{h_{t}\cdot a_{j,t}}{\left\lVert a_{j,t}\right\rVert^{2}}a_{j,t},\quad p_{j,t-1}=h_{t-1}-\frac{h_{t-1}\cdot a_{j,t}}{\left\lVert a_{j,t}\right\rVert^{2}}a_{j,t}
\end{equation}

The normalised vectors of the projection in relation to $s_{j,t}$ are:

\begin{equation}\label{eq:proj_vector}
v_{j,t}=\frac{p_{j,t}- s_{j,t}}{\left\lVert p_{j,t}-s_{j,t}\right\rVert}\quad ,\quad v_{j,t-1}=\frac{p_{j,t-1}- s_{j,t}}{\left\lVert p_{j,t-1}-s_{j,t}\right\rVert}
\end{equation}

The angle between the two vectors, and therefore the change in joint angle is
\begin{equation}\label{eq:theta}
\Delta\theta_{j,t} = \arccos{(v_{j,t-1}\cdot v_{j,t})}\cdot \mathrm{sign}(a_{j,t}\cdot (v_{j,t-1}\times v_{j,t}))
\end{equation}

The new desired joint angle is therefore 

\begin{equation}\label{eq:theta_update}
\theta_{j,t} =
  \begin{cases}
    \theta_{j,t-1} + \mathrm{K}\Delta\theta_{j,t} & \theta_{j,min} \leq \theta_{j,t}  \leq  \theta_{j,max} \\
    \theta_{j,t-1}  & \mathrm{otherwise}
  \end{cases}
\end{equation}

Where $\mathrm{K}$ is a tunable motion scaling factor. \par
This angle update however covers only part of the hand motion, to get the full motion we need to propagate the remaining motion down through the kinematic chain. We rotate $h_{t-1}$ around $a_{j,t}$ centred in $s_{j,t}$  by the angle $(\theta_{j,t}-\theta_{j,t-1})$
\begin{equation}\label{eq:rest}
r_{j,t-1} = q_{j,t}\cdot (h_{t-1}-s_{j,t}) \cdot q_{j,t}^{-1}+s_{j,t}
\end{equation}

Where $q_{j,t}$ is the quaternion representing the rotation around $a_{j,t}$, $q_{j,t}^{-1}$ is it's conjugate, and the $\cdot$ represents the Hamiltonian product. The quaternion $q_{j,t}$ is defined as
\begin{equation}\label{eq:quat}
\begin{gathered} 
\hat{q_{j,t}} = (\cos{(\frac{\theta_{j,t}-\theta_{j,t-1}}{2})},\sin{(\frac{\theta_{j,t}-\theta_{j,t-1}}{2})}a_{j,t}), \\
q_{j,t}=\frac{\hat{q_{j,t}}}{\left\lVert\hat{q_{j,t}}\right\rVert}
\end{gathered}
\end{equation}

\begin{figure}[bh]
    \includegraphics[width=0.48\textwidth]{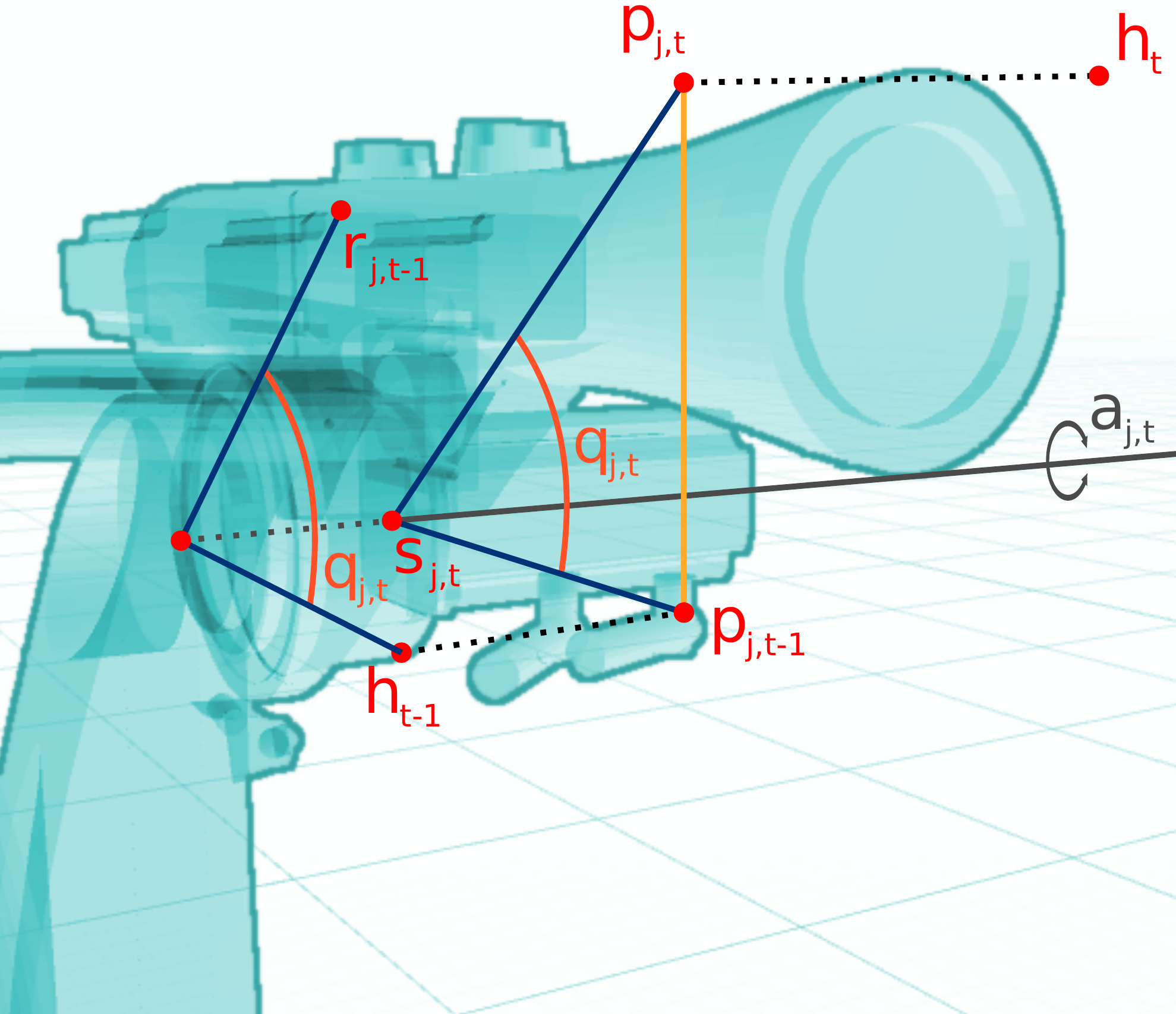}
    \caption{Graphical representation of \eqref{eq:proj} - \eqref{eq:quat}. The hand motion between the two consecutive frames has been vastly exaggerated for the sake of clarity}
    \label{fig:robot_move}
\end{figure}

The same computation is then reiterated for each joint down the kinematic chain until $r_{j,t-1}=h_{t}$ or there are no more joints left. In the latter case the complete motion of the hand is not reproduced by the robot due to kinematic constrains. In Fig.~\ref{fig:robot_move} one can see a graphical representation of the equations. \par

The vector of joint position updates is then sent to ROS where any controller can be selected through ROS Control. If the controller in question is Cartesian, the desired joint values can be converted to the desired end-effector pose via forward kinematics. \par

Given that in the majority of cases dragging the end-effector may be preferable, a holographic sphere is placed around the end-effector which can be dragged and rotated around the three axis of the world coordinate system. The pose of the sphere then indicates the desired pose of the end effector. For control in joint-space the inverse kinematic solution is used that minimizes the total change of all the joint angles. In our case we use the MoveIt! framework to perform both forward and inverse kinematics as needed.\par

\begin{figure}[H]
    \includegraphics[width=0.48\textwidth]{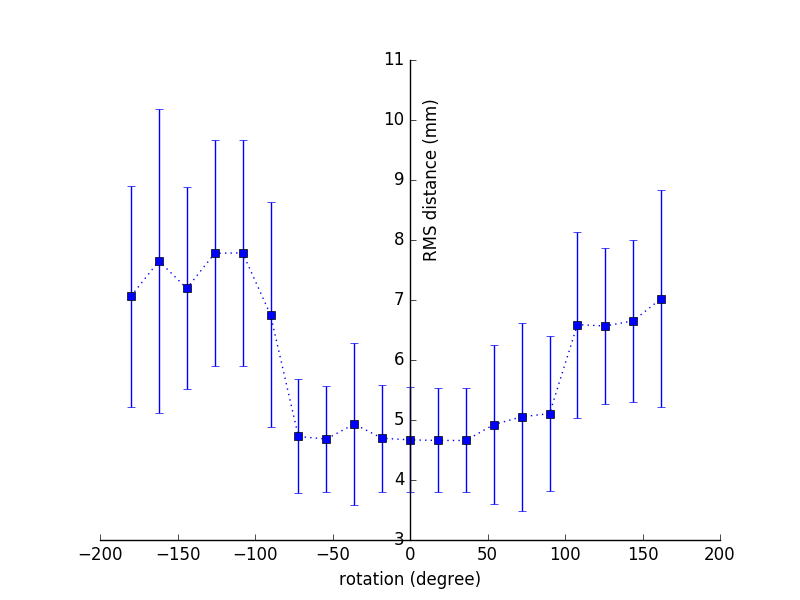}
    \caption{The influence of rotation on our registration algorithm. One can see that the algorithm is robust to rotational errors of the seed hologram, meaning that the user guess doesn't have to be overly precise. One can also see that that error is not mirror due to the robot being asymmetric}
    \label{fig:rotation}
\end{figure}

\section{Experiments and Results}
\label{sec:exp}
To prove the validity of our approach we made two sets of experiments. In the first set, we tested how well does our referencing algorithm perform. This is crucial in the cases of proximal HRI where the robot need to be guided as similarly as possible to JTS-based methods. In the second set of experiments we test how well our control strategy translates hand motion to robot motion. \par

In regards to referencing two main questions needed to be answered. Firstly, is it better than manual referencing and by how much ? Secondly, how precise does the positioning of the seed hologram need to be for the referencing to still be robust? \par

The input point clouds for the registration algorithm were divided into two options, namely a big one (1,240,000 triangles per cubic meter and 256,000 samples per mesh) and a small one (1,000 triangles per cubic meter and 16,000 samples per mesh), with the conversion taking 1 min, and 3 seconds respectively. \par

\begin{table}[H]
\caption{ The mean, minimum, maximum and standard deviation in millimetres of the conducted tests. The first row represents the 12 original human guesses. The ICP-rotation and ICP-translation are the tests conducted by rotating and translating the original user guesses respectively   }
\label{tab:rms}
\begin{tabular}{ |l|l|l|l|l|} 
 \hline
 &mean(mm)&min(mm)&max(mm)&$\sigma$ (mm)\\ 
 \hline
 User guess & 27.53 & 3.65 & 138.87 & 44.52 \\ 
 ICP - all & 4.91 & 3.22 & 14.42 & 1.39 \\ 
 ICP - rotation & 5.90 &  3.22 & 14.21 & 1.90 \\
 ICP - translation & 4.74 & 3.22 & 14.41 & 1.20 \\
 \hline
\end{tabular}
\end{table}

\begin{figure*}[t]
\centering
    \includegraphics[width=0.45\textwidth]{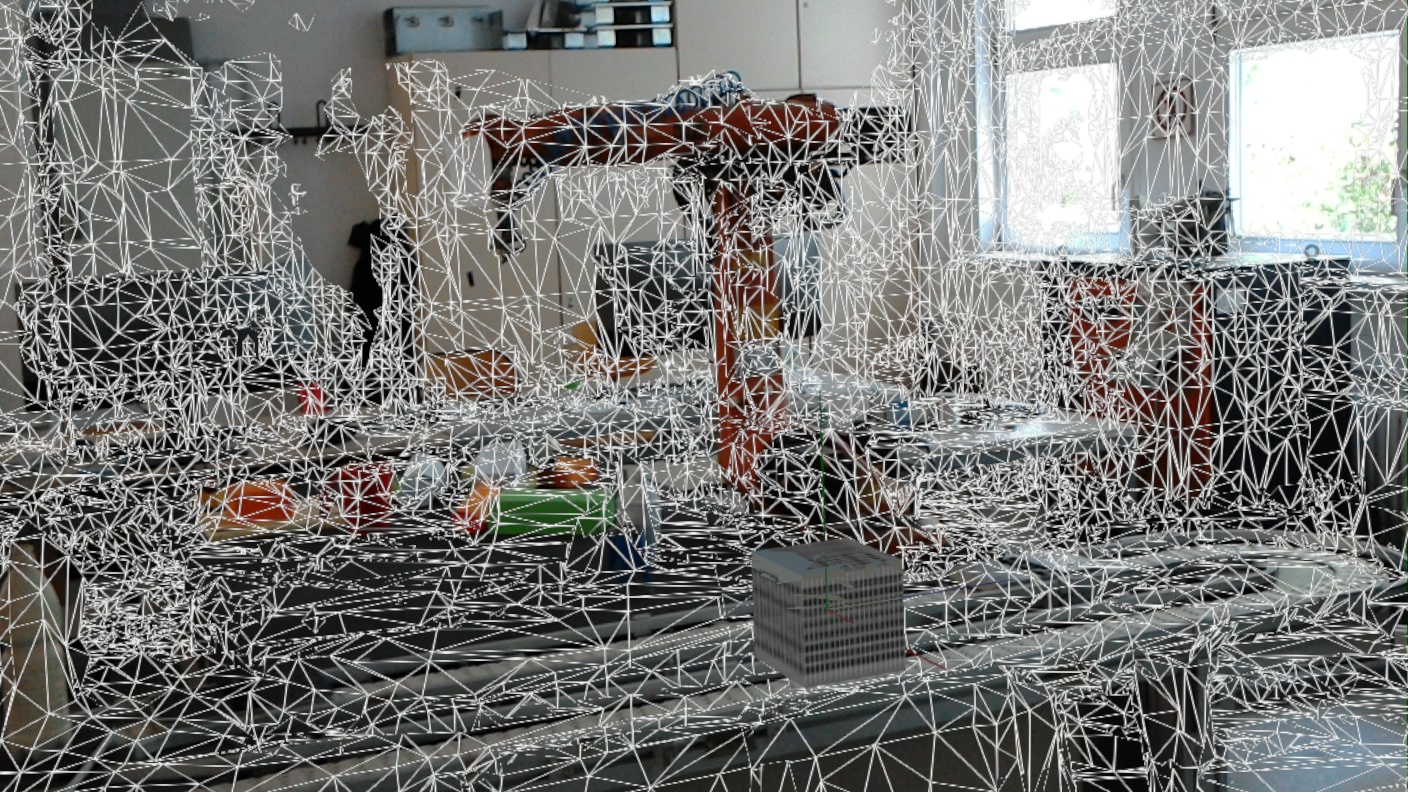}
    \includegraphics[width=0.45\textwidth]{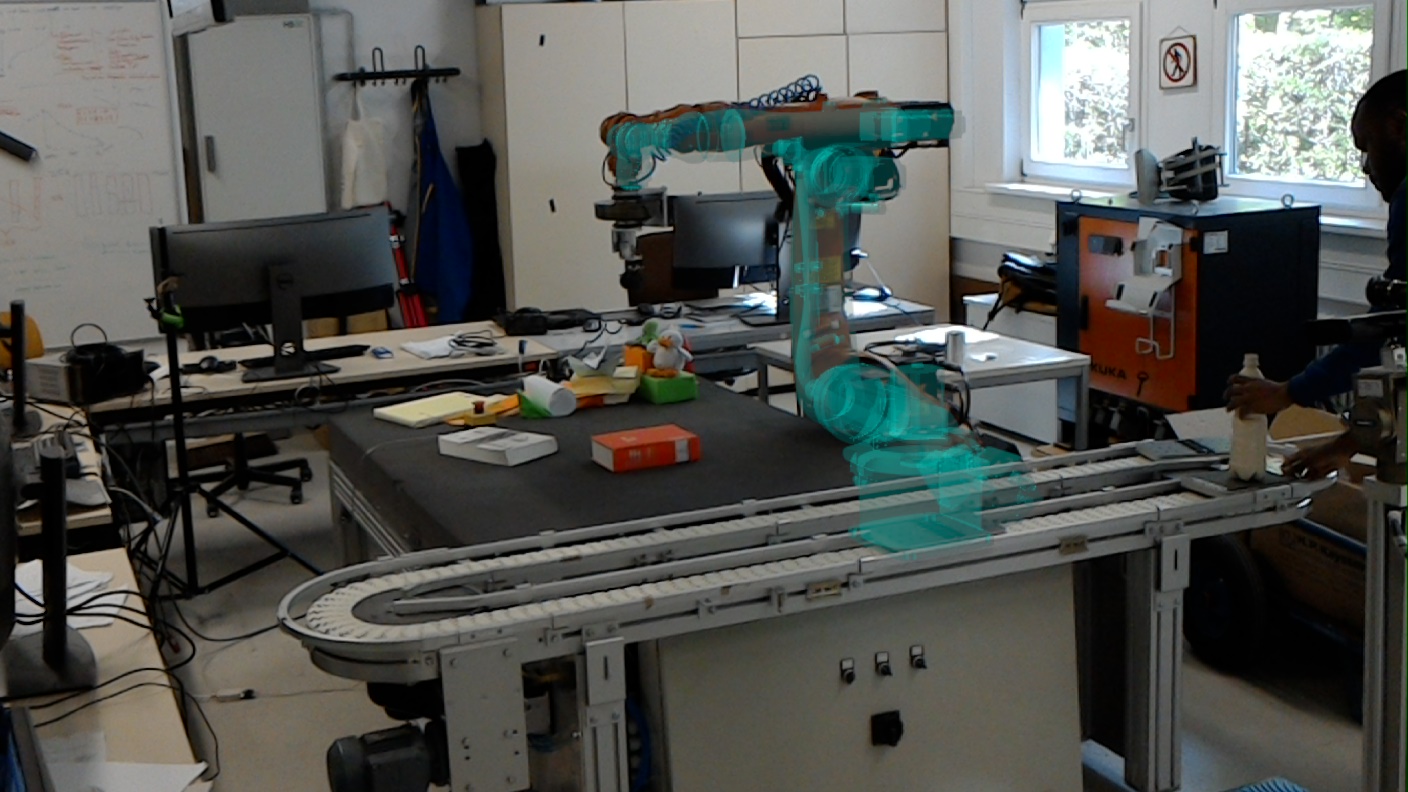}
    \caption[]{The results of the referencing algorithm; (left) The HoloLens spatial mesh and the seed hologram positioned by the user; (right) The hologram overlayed on the real robot}
    \label{fig:ref_result}
\end{figure*}

Preliminary experiments showed that in our test scenario Super4PCS had poor performance. The reason for this being that it requires segmented out models from the scene to work, being a global descriptor algorithm. Our test cell has a table very close to the robot, as seen in Fig.~\ref{fig:intro} which was not possible to be segmented-out fully. Furthermore in experiments with a similar model, the KR-6, it showed very wide precision deviation based on the parameters chosen. ICP on the other hand proved much more robust in case of parameter changes. The small cloud was chosen as it had comparable performance with much less processing time.  \par

The root mean square (RMS) distance between the closest points in the two point clouds was used as a metric for the precision of the referencing. As a second evaluation, the matches were visually inspected, as some bad matches could have quite low RMS distances if close to other objects in the scene. These results were disregarded. Twelve user guesses and point clouds were taken. Each user guess was then rotated in steps of 18\textdegree{} to test the influence of imprecise rotation of the seed hologram. The influence of rotation on the ICP can be seen in Fig.~\ref{fig:rotation}. Keeping the original rotation, the seed algorithm was translated in a 1m volume around the initial guess with a step of 0.1m. In Table ~\ref{tab:rms} one can see the precision of original user guesses, the ICP subject to rotation of the seed algorithms, the ICP subject to translation, and the average statistics of all cases. One can see that the ICP performs much better and with less deviation than a purely manual method. The referencing result can be seen in Fig.~\ref{fig:ref_result}.
\par

\begin{figure*}[h]
\centering
    \includegraphics[width=0.45\textwidth]{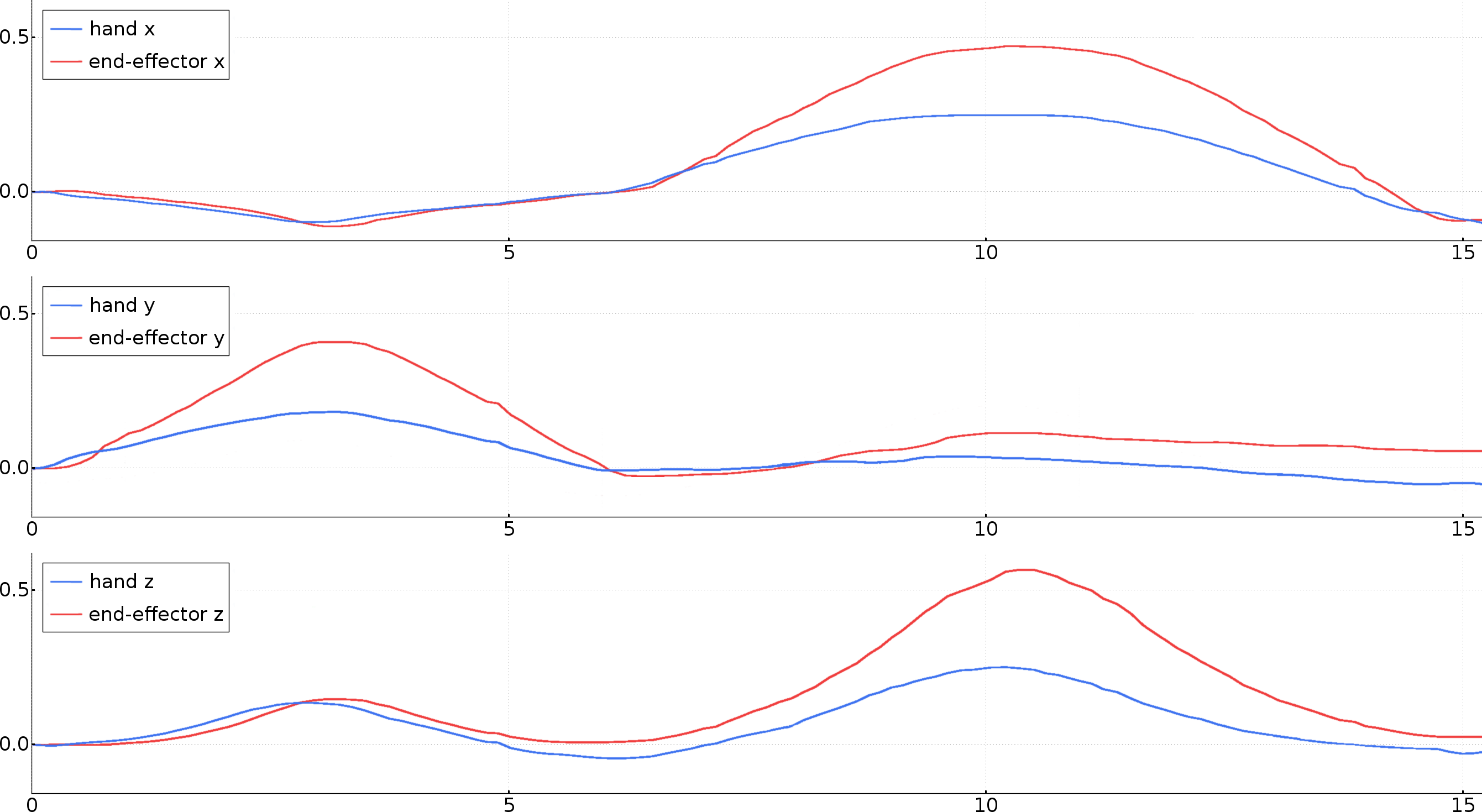}
    \includegraphics[width=0.45\textwidth]{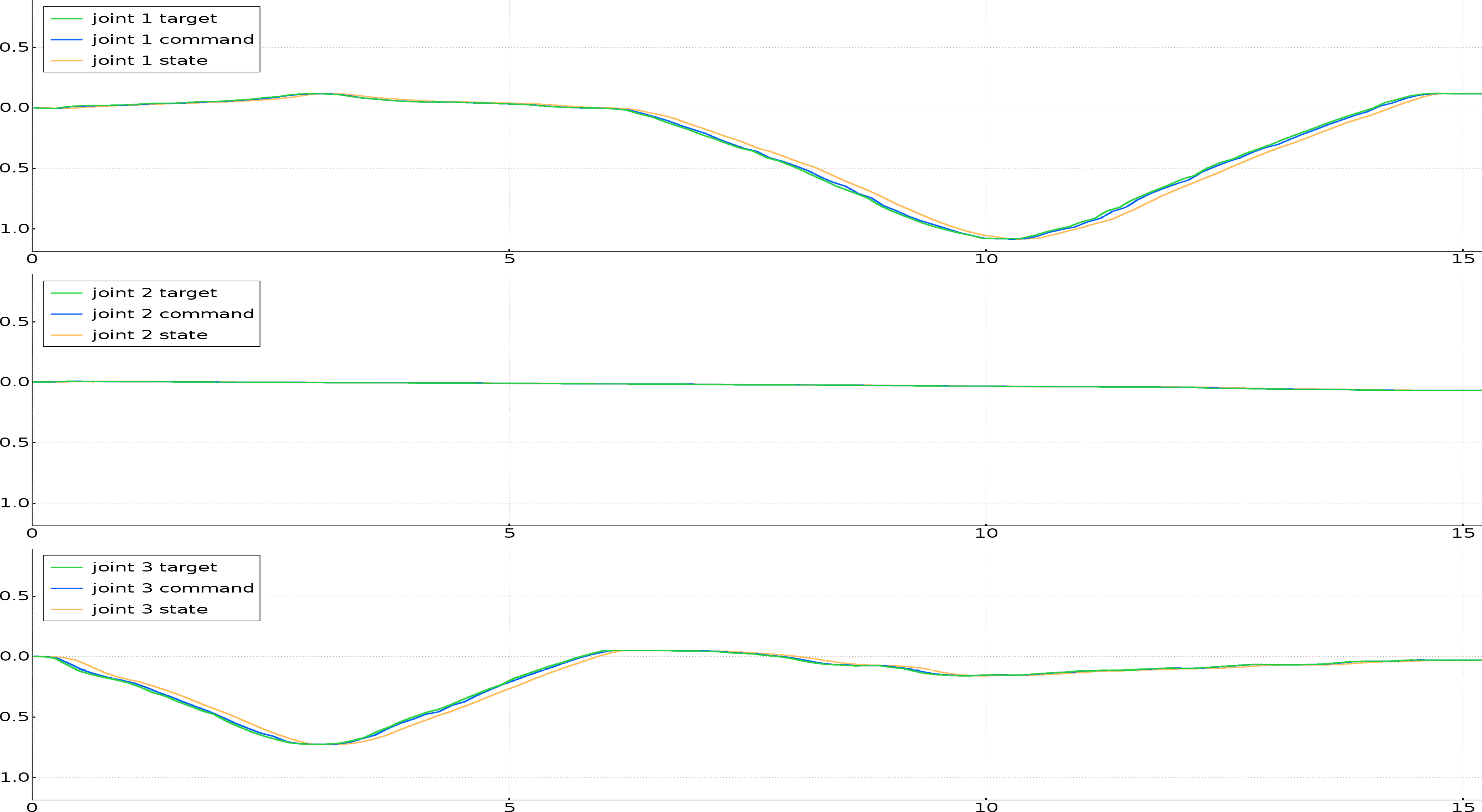}
    \caption[]{The Cartesian movements of the end-effector and joints 1-3 while moving the third link for 15 seconds. (left) The movements of the hand strongly correlate with those of the end-effector and therefore link 3, as the three joints up the kinematic chain weren't changed. The movements of the end-effector are larger than that of the hand as the end-effector is further away from the third joint. The y axis is in meters and the x axis in seconds (right) The joint commands of joints 1-3, with joint 3 being the closest to link 3. As the axis of rotations of joints two and three are parallel, one can see almost no change in joint two. One can also see the state of the robot lags behind the commands due to inertia. The y axis is in radians and the x in seconds}
    \label{fig:graphs}
\end{figure*}

Tests of the the control strategy were performed using $\mathrm{K}=1$ and a Reflexxes interpolated joint position controller \cite{Kroeger2011reflexxes}. Moving each link, we compared the positions of the end-effector to the hand position, as well as tracked the desired joint state provided by our algorithm, the control signal of the Reflexxes controller, and the actual joint state of the robot. The robot used was an industrial KUKA KR-5 ARC manipulator with no JTS. In Fig.~\ref{fig:graphs} one can see the example where link 3 is moved. The end-effector, and therefore link 3, motions follow closely the hand motions. One can also see that, although the hand commands suffer from shaking as they were unfiltered, the inertia of the robot filters out this signal and the joint states are smoother than the commands. That said a tunable band-stop filter, or signal smoothing for small, shaky hand motions could be a helpful addition.\par

Also visible in Fig.~\ref{fig:graphs}, the desired joint values follow closely what one would intuitively expect from our command strategy, with the base joint (joint 1) and elbow joint (joint3) receiving the majority of move commands. The non-zero move command from the shoulder joint (joint 2) is due to the fact that when $s_{3,t}$, $h_{t}$ and $h_{t-1}$ are collinear, the projection angle is zero, and therefore the next joint, joint 2, needs to perform the movement. \par 


\section{Conclusions and Future Work}
\label{sec:conc}
To extended the option of using hand guidance to any robot possessing an urdf description and a 3D model, we developed a system that uses ICP-based registration to find a robust transform between the Microsoft HoloLens headset and the robot. After that the user has the option to perform either telemanipulation or direct hand guidance near the robot, either through end-effector dragging, or through pushing and pulling specific links which is very useful for robots with redundant kinematics. We proved the feasibility of this approach by testing both the robustness of the referencing strategy and the control strategy using a KR-5 industrial manipulator. The referencing performed better than a pure human guess, with good resilience to errors in translation and rotation of the seed hologram. Meanwhile the control strategy successfully moved the robot in a way that closely follows the motions of the user's hand.  \par   

The in-built hand-tracking used in this paper had difficulties tracking the hand in front of black backgrounds, likely due to the absorption properties, and when in close proximity to objects, due to the fact that the hand can't be segmented out from the background. To alleviate these problems, future implementations will transition to the method proposed in \cite{GANeratedHands_CVPR2018}, which can deal both with contact cases and a dark background as it uses RGB images instead of an infrared depth stream. Furthermore it can track the hand of the user in any pose, meaning that the user's arm doesn't need to adhere to the few pre-programmed gestures of the in-built hand tracking. \par

Future work also includes tests with more robots to detect any failure cases of either referencing or the control strategy, measuring the referencing error with respect to the ground truth, and finally a user study where our approach is compared with one based on JTSs to evaluate the general usability. \par


\section*{ACKNOWLEDGMENT}
This work has been supported from the European Union’s Horizon 2020 research and innovation programme under grant agreement No 688117 “Safe human-robot interaction in logistic applications for highly flexible warehouses (SafeLog)”.

\bibliographystyle{IEEEtran}
\bibliography{bibliography}

\end{document}